\documentclass{article}
\usepackage{spconf,amsmath,graphicx}
\usepackage{enumerate}

\newcommand{\tabincell}[2]{\begin{tabular}{@{}#1@{}}#2\end{tabular}}
\title{Automatic Prosody Prediction for Chinese Speech Synthesis using BLSTM-RNN and Embedding Features}

\name{Chuang Ding$^1$, Lei Xie$^{1, 2}$, Jie Yan$^2$, Weini Zhang$^2$, Yang Liu$^2$}
\address{
  $^1$School of Computer Science, Northwestern Polytechnical University, Xi'an, China \\
  $^2$School of Software and Microelectronics, Northwestern Polytechnical University, Xi'an, China \\
  {\tt \{cding, lxie, jyan, wnzhang, yangliu\}@nwpu-aslp.org}
}

\begin{document}
%\ninept
%
\maketitle
\begin{abstract}
Prosody affects the naturalness and intelligibility of speech. However, automatic prosody prediction from text for Chinese speech synthesis is still a great challenge and the traditional conditional random fields (CRF) based method always heavily relies on feature engineering. In this paper, we propose to use neural networks to predict prosodic boundary labels directly from Chinese characters without any feature engineering. Experimental results show that stacking feed-forward and bidirectional long short-term memory (BLSTM) recurrent network layers achieves superior performance over the CRF-based method. The embedding features learned from raw text further enhance the performance.
\end{abstract}
\begin{keywords}
automatic prosody prediction, speech synthesis, neural network, BLSTM, embedding features
\end{keywords}

\vspace{-6pt}
\section{Introduction}

\begin{figure}[t]
  \centering
  \vspace{-10pt}
  \includegraphics[width=70 mm]{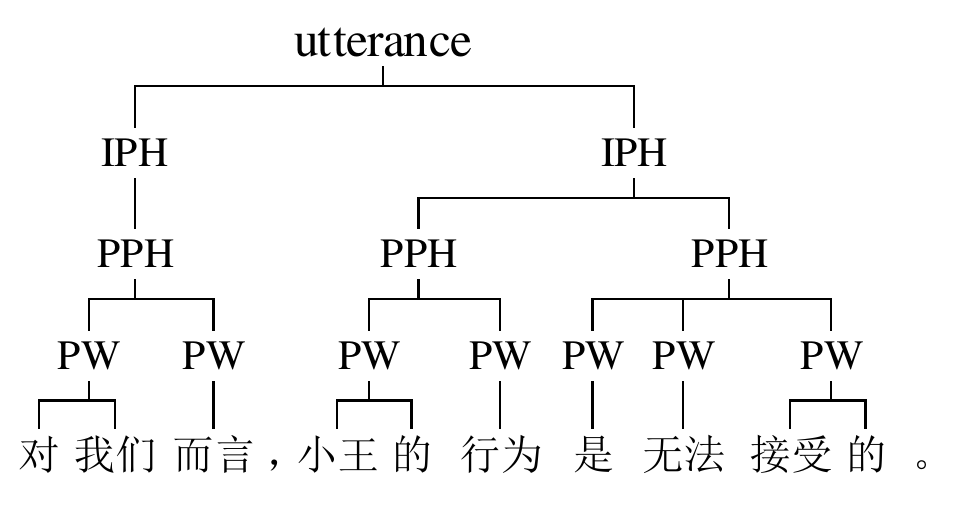}
  \vspace{-10pt}
  \caption{Three-layer prosody structure tree in Chinese.}
  \vspace{-12pt}
  \label{fig:prosody-structure}
\end{figure}

Prosody refers to the rhythm, stress and intonation of speech, including variations in duration, loudness and pitch. It is well known that speech prosody plays an important perceptual role in human speech communication~\cite{qian2010automatic}. Specifically, perception of prosodic boundaries is essential for listeners. In Chinese speech synthesis systems, typical prosody boundary labels consist of prosodic word (PW), prosodic phrase (PPH) and intonational phrase (IPH), which construct a three-layer prosody structure tree~\cite{sun2009chinese}, as shown in Fig.~\ref{fig:prosody-structure}. The leaf nodes of tree structure are lexical words that can be derived from a lexical-based word segmentation module. Whether the prosody labels are properly predicted will directly affect the naturalness and intelligibility of the synthesized speech.

Previous studies have investigated a great number of features, their relevance to prosody generation in speech production and various prosodic modeling methods. Some syntactic cues like part-of-speech (POS), syllable identity, syllable stress and their contextual counterparts are commonly used for prosody boundary prediction~\cite{jeon2009automatic,rangarajan2007exploiting,koehn2000improving}. Many statistical methods have been investigated to model speech prosody, including classification and regression tree~\cite{chu2001locating}, hidden Markov model~\cite{nie2003automatic}, maximum entropy model~\cite{li2004chinese} and conditional random fields (CRF)~\cite{levow2008automatic}. To our knowledge, the best reported results were achieved with CRF due to its ability of relaxing strong model independence assumption and solving the label bias problem~\cite{qian2010automatic,LaffertyCRF}.

Despite years of research, it is still a great challenge to predict correct prosodic labels from unrestricted text for a text-to-speech (TTS) system. Obviously, there are two major drawbacks of the CRF-based prosody prediction in Chinese speech synthesis. First, it heavily relies on the performances of Chinese word segmentation (CWS) and POS tagging~\cite{sheng2002learning}. Second, the particle size and the inevitable segmentation errors in CWS have negative effects on the subsequent prosodic boundary prediction task. Moreover, the choice of effective features, from a broad set of feature templates, is critical to the success of such systems~\cite{zheng2013deep}. Much of the effort goes into feature engineering, which is notoriously labor-intensive, mainly based on the experience of an annotator.

Recently, deep neural networks (DNN) have been increasingly investigated in order to minimize the effort of feature engineering in sequential labeling tasks. Zheng et al.~\cite{zheng2013deep} applied neural networks to CWS and POS tagging and proposed a perceptron-style algorithm to speed up the training process with negligible loss in performance. Pei et al.~\cite{pei2014maxmargin} proposed a max-margin tensor neural network for CWS to model interactions between tags and context characters by exploiting tag embeddings and tensor-based transformation. These researches have proved that DNN is able to achieve similar or even superior performance over CRF-based method with minimal feature engineering in sequential labeling tasks. Therefore, it is promising to apply DNN architectures to automatic prosody prediction. However, we notice that the neural networks used in previous researches are feed-forward structures that keep the assumption of sample independence and provide only limited context modeling ability by operating on a fixed-size window of input samples. Instead, bidirectional recurrent neural networks (BRNN) are able to incorporate contextual information from both past and future inputs~\cite{schuster1997bidirectional}. Specifically, BRNN with long short-term memory (LSTM) cells, namely BLSTM-RNN, has become a popular model~\cite{hochreiter1997long}.

In this paper, we address the prosodic boundary prediction problem using neural networks. There are three main contributions. (1) We propose a neural network approach to predict prosody labels directly from Chinese characters without any feature engineering. (2) We show that superior performance is achieved by stacking feed-forward and bidirectional long short-term memory (BLSTM) recurrent layers. (3) We leverage a large raw text corpus to obtain useful character embedding features. Both objective and subjective evaluations show that the proposed architecture achieves superior performance over the CRF-based method and the embedding features further enhance the performance.

\vspace{-10pt}
\section{The Proposed Approach}
\begin{figure}[t]
  \centering
  \vspace{-10pt}
  \includegraphics[width=70 mm]{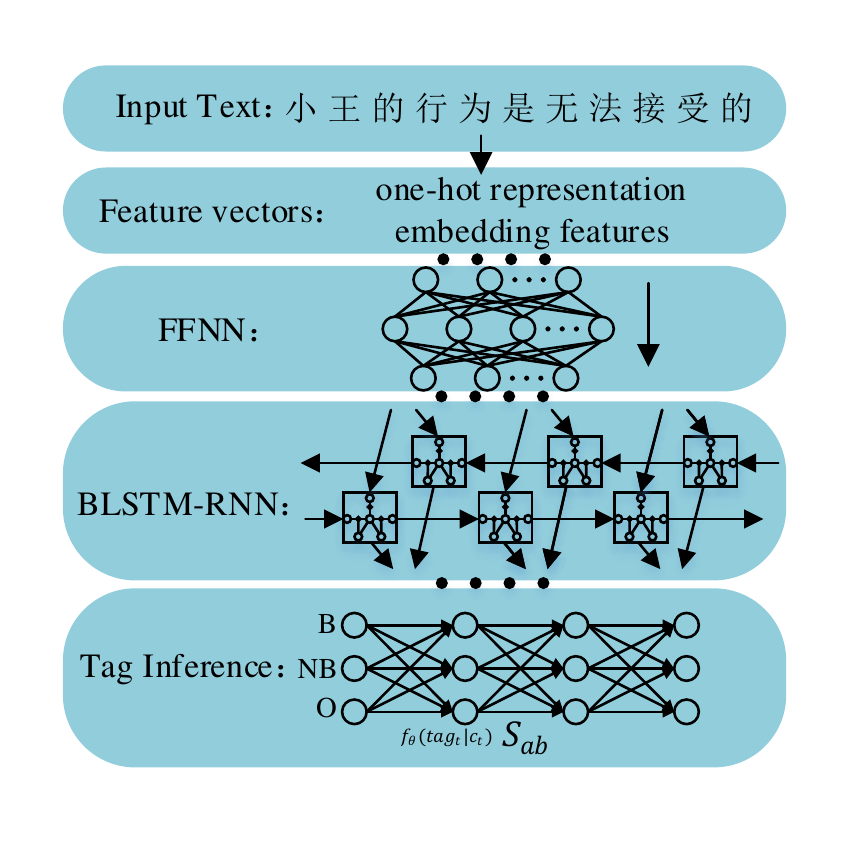}
  \vspace{-23pt}
  \caption{The neural network architecture for prosodic boundary prediction. In tag inference, B, NB and O denote boundary, non-boundary and others (e.g., punctuation), respectively.}
  \vspace{-12pt}
  \label{fig:NNarchitecture}
\end{figure}

\vspace{-3pt}
Just like CWS and POS tagging, automatic prosody prediction can be treated as a sequential labeling task that assigns boundary labels to characters of an input sentence. In order to make the prediction models less dependent on the feature engineering, we choose to use a variant of the neural network architecture proposed by~\cite{bengio2003neural} for probabilistic language model. This architecture was subsequently used for CWS and POS tagging~\cite{zheng2013deep}.  As shown in Fig.~\ref{fig:NNarchitecture}, the architecture takes raw text as input and maps each Chinese character into a basic feature vector. The following layers are two types of neural networks, FFNN and BLSTM-RNN, used to discover multiple levels of feature representations from the basic feature vectors. The output layer is a graph over which tag inference is achieved by the Viterbi algorithm.

%\vspace{-5pt}
%\section{Network Training}

\vspace{-6pt}
\subsection{Feature Vectors}
The characters fed into network are transformed into feature vectors by a mapping operation. Typically, a character dictionary $D$ of size $|D|$ is extracted from the training set and unknown characters are mapped to a special symbol that is not used elsewhere. Each Chinese character can be typically represented by a one-hot vector, the size of which is $|D|$, and all dimensions are marked as $0$ except the location of the character in $D$, which is marked as $1$.  However, the one-hot representation, with high dimensions, fails to model the semantic similarity between the ideographic characters. In contrast, the \textit{distributed} representation or \textit{embedding} feature, in form of a low dimensional
continuous-valued vector learned using neural networks from raw text in a fully unsupervised way, is assumed to carry important syntactic and semantic information~\cite{mikolov2013distributed}~\cite{mikolov2013linguistic}. Recently, Mansur et al.~\cite{mansur2013feature} have shown superior performance in Chinese word segmentation by the use of embedding features based on a neural language model~\cite{bengio2003neural}. Besides ~\cite{bengio2003neural}, Mikolov et al.~\cite{mikolov2013distributed} proposed a faster skip-gram model called \emph{word2vec\footnote{https://code.google.com/p/word2vec/}}. As our preliminary experiments do not show much performance difference among various embedding features, we simply choose~\emph{word2vec} in this study because it can be trained much faster.

\vspace{-10pt}
\subsection{Network Structures and Training}
Two types of neural networks are investigated in this paper: FFNN and BLSTM-RNN. FFNN, trained with a back-propagation learning algorithm~\cite{horikawa1992fuzzy}, is widely used in many practical applications. In a typical FFNN, every unit in a layer is connected with all the units in the previous layer, which takes in the output of the previous layer and computes a new set of non-linear activations for next layer. However, the assumption of sample independence brings in only limited context modeling ability.

\begin{figure}[t]
  \centering
  \vspace{-10pt}
  \includegraphics[width=80 mm]{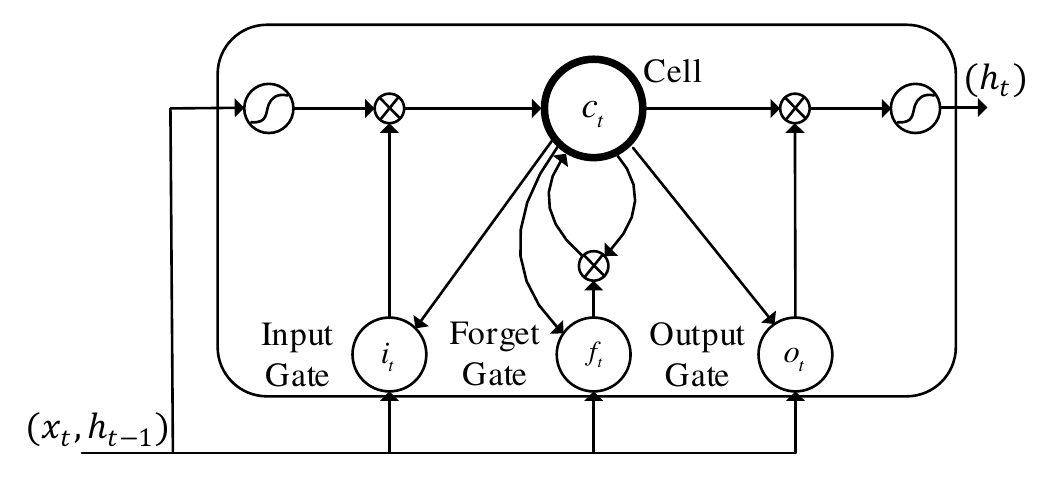}
  \vspace{-10pt}
  \caption{Long short-term memory cell.}
  \vspace{-10pt}
  \label{fig:LSTMcell}
\end{figure}

Researchers have proposed RNN to solve the limitation of FFNN. However, conventional RNN is only able to make use of previous context information. This is not accurate in modeling speech prosody that is highly related with both past and future contexts. Instead, bidirectional RNN can access both the preceding and succeeding input contexts with two separate hidden layers, which are then fed to the same output layer. The activation function $\mathcal{H}$ of RNN is usually a sigmoid or hyperbolic tangent function, which often causes the gradient vanishing problem that prevents RNN from modeling the long-span relations in sequence features. An LSTM architecture, which uses purpose-built memory cells to store information, can overcome this problem and model longer contexts. Fig.~\ref{fig:LSTMcell} illustrates a single LSTM memory cell. For LSTM, $\mathcal{H}$ is implemented by the following functions:
\begin{align*}
& i_t = \sigma(W_{xi}x_t + W_{hi}h_{t-1} + W_{ci}c_{t-1} + b_i) \\
\end{align*}

\begin{align*}
& f_t = \sigma(W_{xf}x_t + W_{hf}h_{t-1} + W_{cf}c_{t-1} + b_f) \\
& c_t = f_tc_{t-1} + i_ttanh(W_{xc}x_t + W_{hc}h_{t-1} + b_c) \\
& o_t = \sigma(W_{xo}x_t + W_{ho}h_{t-1} + W_{co}c_t + b_o) \\
& h_t = o_ttanh(c_t)
\end{align*}
where $x = (x_1, x_2, ... x_t ..., x_T)$ is the input feature sequence, $\sigma$ is the logistic function, and $i$, $f$, $o$ and $c$ are the input gate, forget gate, output gate and cell memory, respectively. $W$ is the weight matrix and the subscript indicates it is the matrix between two different gates.

BLSTM-RNN is a combination of LSTM and BRNN. Deep bidirectional LSTM-RNN can be established by stacking multiple BLSTM-RNN hidden layers on top of each other. The output sequence of one layer is used as the input sequence of the next layer. The hidden state sequences, $h^n$, consist of forward and backward sequences $\overrightarrow{h}^n$ and $\overleftarrow{h}^n$, iteratively computed from $n = 1$ to $N$ and $t = 1$ to $T$ as follows:
\begin{align*}
& \overrightarrow{h}_t^n = \mathcal{H}(W_{\overrightarrow{h}^{n-1}\overrightarrow{h}^n}\overrightarrow{h}_t^{n-1} + W_{\overrightarrow{h}^n\overrightarrow{h}^n}\overrightarrow{h}_{t - 1}^n + b_{\overrightarrow{h}}^n), \\
& \overleftarrow{h}_t^n = \mathcal{H}(W_{\overleftarrow{h}^{n-1}\overleftarrow{h}^n}\overleftarrow{h}_t^{n-1} + W_{\overleftarrow{h}^n\overleftarrow{h}^n}\overleftarrow{h}_{t - 1}^n + b_{\overleftarrow{h}}^n), \\
& y_t = W_{\overrightarrow{h}^Ny}\overrightarrow{h}_t^N + W_{\overleftarrow{h}^Ny}\overleftarrow{h}_t^N + b_y.
\end{align*}
where  $y = (y_1, y_2, ... y_t ..., y_T)$ is the output prosodic boundary sequence.

In our study, the feed-forward layers are trained with typical backpropagation (BP) algorithm and the back-propagation through time (BPTT) method is used for training of BLSTM layers. BPTT is applied to both forward and backward hidden nodes and back-propagates layer by layer. The weight gradients are computed over the entire utterance~\cite{williams1995gradient}. The neural networks can be trained effectively in a layer-wised training manner, which makes it convenient to stack different types of neural network layers on top of each other to form a deep architecture. The deep architecture is able to build up progressively higher level representations of the input data, which is a crucial factor of the recent success of hybrid systems~\cite{graves2013hybrid}.

%\begin{figure}[t]
%  \centering
%  \vspace{-10pt}
%  \includegraphics[width=90 mm]{figure/blstm-rnn.pdf}
%  \vspace{-20pt}
%  \caption{Bidirectional LSTM-RNN layer.}
%  \vspace{-13pt}
%  \label{fig:blstm-rnn}
%\end{figure}

\subsection{Tag Inference}

To model the tag dependency and infer the tag sequence globally, given a set of tags $G=\{B, NB, O\}$, a transition score $S_{ab}$ is introduced for jumping from tag $a \in G$ to tag $b \in G$. For the input character sequence of a sentence $c_{[1:T]}$ with a tag sequence $tag_{[1:T]}$, a sentence-level score is then given by the sum of transition and network scores~\cite{zheng2013deep,collobert2011natural}:
\vspace{-10pt}
\begin{align*}
& l(c_{[1:T]},tag_{[1:T]},\theta) = \sum_{t=1}^{T}(S_{tag_{t-1}tag_t} + f_\theta(tag_t|c_t))
\end{align*}
where $f_\theta(tag_t|c_t))$ indicates the score output for $tag_t$ at the $t$-th character by the networks. Given a sentence $c_{[1:T]}$, we can find the best tag path $tag^*_{[1:T]}$ by maximizing the sentence score:
\vspace{-10pt}
\begin{align*}
& tag^*_{[1:T]} = \arg\max_{\forall l_{[1:T]}} l(c_{[1:T]}, tag_{[1:T]}, \theta).
\end{align*}
The Viterbi algorithm can be used for tag inference. The description above shows that it is easy to stack feature vectors, neural networks and tag inference together. Thus, the proposed architecture can be trained in a layer-wised fashion.

%The system in our experiments is altered on CURRENNT, a machine learning library for FFNN and LSTM-RNN which uses GPU to accelerate the computations~\cite{weninger2014introducing}.

\section{Experiments}

Totally 48210 sentences randomly selected from People's Daily were used in our experiments. Prosodic boundaries (PW, PPH and IPH) were labelled by professional annotators with corresponding speech and labeling consistency is ensured. Word segmentation and POS tagging were carried out by a front-end preprocessing tool. The accuracy of word segmentation is 97\% and the accuracy of POS tagging is 96\%. The corpus was partitioned into three parts: a training set with 43390 utterances, a validation set with 2410 utterances for parameter tuning and a testing set with another 2410 utterances. A character dictionary $D$ of size 4030 was extracted from the training set. A large set of raw texts was also collected from People's Daily for unsupervised embedding feature learning. All texts were preprocessed with text normalization.

In the experiments, PW, PPH and IPH were predicted separately. That is to say, three separate neural network models were trained independently for PW, PPH and IPH using the CURRENNT toolkit~\cite{weninger2014introducing}. Each character in a sentence was assigned to one of the following three boundary tags: B for a prosodic boundary, NB for a non-boundary, and O for other symbols such as punctuation. Precision (P), recall (R) and F-score (F) were calculated as standard objective evaluation criteria.

A CRF-based prosodic boundary prediction approach was used as baseline and boundary prediction (B, NB and O) was operated at word level. Atomic features in the CRF approach include word identity, POS tags, the length of word and the predicted tag from the previous boundary level. A linear statistical model was applied to optimize the feature templates. Parameters grid search was adopted to achieve the best performance of the CRF model. The CRF++ toolkit\footnote{http://taku910.github.io/crfpp/} was used for the CRF-based prosodic boundary prediction. The baseline results are shown in Table~\ref{tab:resultCRF}.

\begin{table}[t]
  \small
  \centering
  \vspace{-10pt}
  \begin{tabular}{|c|c|c|c|}
    \hline
    Boundary & P (\%) & R (\%) & F (\%) \\ \hline \hline
    PW & 95.34 & 96.73 & 96.03 \\ \hline
    PPH & 83.41 & 83.68 & 83.06 \\ \hline
    IPH & 84.85 & 73.39 & 78.71 \\ \hline
  \end{tabular}
  \caption{The results of CRF-based prosody prediction.}
  \label{tab:resultCRF}
\end{table}

\begin{table}[t]
  \small
  \centering
  \begin{tabular}{|c|c|}
    \hline
    Topology & \tabincell{c}{ B, BB, BBB, BBBB \\ FFB, FBF, BFF, FBB, BFB, BBF }\\ \hline
    Num of nodes & 32, 64, 128, 256 \\ \hline
  \end{tabular}
  \caption{Different network configurations in the experiments.}
  \vspace{-10pt}
  \label{tab:networkcnf}
\end{table}

We investigated the performance of neural network architecture with different topologies, as described in Table~\ref{tab:networkcnf}, where F and B denote a feed-forward layer and a BLSTM layer, respectively. The number of the nodes were kept the same for all hidden layers in every tested network architecture. Specifically, the network input is an $M$-dimensional feature vector, where $M$=4030 for the PW prediction and $M$=4031 for the PPH and IPH prediction~\footnote{The predicted tag from the previous level was used as a feature.}. The network output corresponds to the three boundary tags (B, NB and O). All networks were trained with a momentum of 0.9, a learning rate of 1e-3 for PW and 1e-4 for PPH and IPH. BPTT was performed using stochastic gradient descent (SGD) with 32 parallel sentences. The training stops if no lower error on the validation set can be achieved within the last 10 epochs. The best performances for different prosodic boundary levels are shown in Table~\ref{tab:fbb}. We interestingly discover that the best performances at different levels are all obtained with a topology of FBB. When we compare Table 3 with the CRF-baseline Table 1, we find that the proposed neural network approach achieves competitive performance at the PW level and significant improvements at the PPH and IPH levels.

We also studied the effectiveness of the character embedding features. Different sizes of unsupervised training data (400M, 800M, 1200M, 1600M and 2000M text) and embedding feature sizes (100, 200, 300 and 400) were tested. The best network architectures, as shown in Table 3, were used in the experiments. Please note that the dimension of feature vector is greatly reduced as compared with the one-hot representation. The results shown in Table~\ref{tab:embedding} indicate that the embedding features can further improve the performance of automatic prosodic boundary prediction.

\begin{table}[t]
  \small
  \centering
  \vspace{-10pt}
  \begin{tabular}{|c|c|c|c|c|}
    \hline
    Boundary & P (\%) & R (\%) & F (\%) & TP / Num of nodes \\ \hline \hline
    PW & 96.02 & 96.69 & 96.35 & FBB / 32 \\ \hline
    PPH & 82.50 & 86.75 & 84.57 & FBB / 128 \\ \hline
    IPH & 84.06 & 79.33 & 81.63 & FBB / 64 \\ \hline
  \end{tabular}\vspace{-2pt}
  \caption{The best performance of each level and the corresponding network topology (TP).}
  \label{tab:fbb}
\end{table}

\begin{table}
  \small
  \centering
  \begin{tabular}{|c|c|c|c|c|}
    \hline
    Boundary & P (\%) & R (\%) & F (\%) & Embedding feature size \\ \hline \hline
    PW & 96.27 & 96.91 & 96.59 & 300 \\ \hline
    PPH & 82.89 & 87.13 & 84.96 & 400  \\ \hline
    IPH & 84.81 & 79.88 & 82.27 & 100 \\ \hline
  \end{tabular}
  \caption{The results of neural network architecture with embedding features and the corresponding feature size.}
  \vspace{-10pt}
  \label{tab:embedding}
\end{table}

We further conducted an A/B preference test on the naturalness of the synthesized speech. A set of 100 sentences were randomly selected from the test set and the prosodic boundary labels were achieved by:
\begin{itemize}
\item[(1)] CRF-based model in Table~\ref{tab:resultCRF};
\item[(2)] NN with one-hot representation in Table~\ref{tab:fbb};
\item[(3)] NN with embedding features in Table~\ref{tab:embedding}.
\end{itemize}

\begin{figure}[htbp]
  \centering
  \vspace{-15pt}
  \includegraphics[width=90 mm]{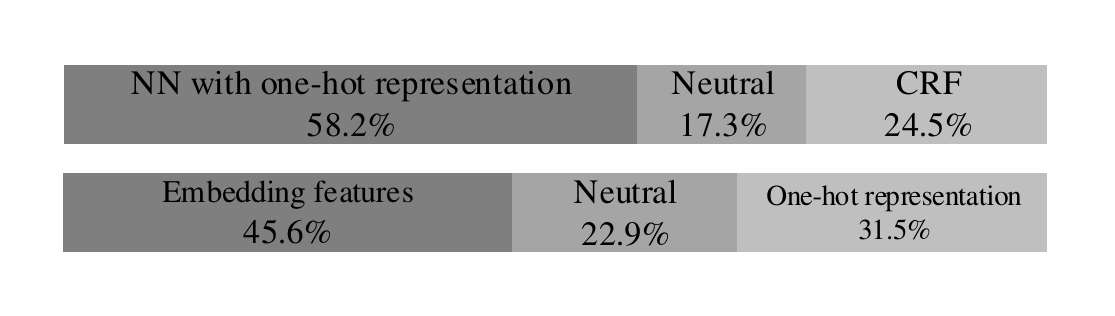}
  \vspace{-25pt}
  \caption{The percentage preference of A/B test.}
  \label{fig:sub}
\end{figure}

We carried out two sessions of comparative evaluations: (1) vs (2) and (2) vs (3). A set of 20 sentence pairs of each session was randomly selected from the 100 pairs with different prosody prediction results and speech was generated through a typical HMM-based TTS system. A group of 10 subjects were asked to choose which one was better in terms of the naturalness of synthesis speech. The percentage preference is shown in Figure~\ref{fig:sub}. We can clearly see that the NN architecture with one-hot representation can achieve better naturalness of synthesized speech as compared with CRF, while the use of embedding features further improves the natrualness.

\vspace{-10pt}

\section{Conclusion and Future Work}
In this paper, we propose to use neural network architectures to predict prosodic boundary labels directly from Chinese characters without feature engineering. We show that superior performance is achieved by stacking feed-forward and bidirectional long short-term memory (BLSTM) recurrent layers. We obtain useful character embedding features from raw text. Both objective and subjective evaluations show that the proposed neural network approch achieves superior performance over the CRF-based approach and the use of embedding features can further boost the performance. For future work, it is promising to predict PW, PPH and IPH labels in a unified neural network and n-gram character embedding features can be further investigated.

\section{Acknowledgements}
This work was supported by the National Natural Science Foundation of China (61175018 and 61571363).

\bibliographystyle{IEEEbib}
\bibliography{refs}

\end{document}